\documentclass[a4paper, 10pt]{IEEEtran}  

\normalfont\selectfont
\usepackage{geometry}
\usepackage[square,numbers]{natbib}
\usepackage{graphics} 
\usepackage{epsfig} 
\usepackage{mathptmx} 
\usepackage{times} 
\usepackage{amsmath} 
\usepackage{amssymb}  
\usepackage{blindtext}
\usepackage{graphicx}
\usepackage{caption}
\usepackage{color} 
\usepackage{array,multirow}
\usepackage{gensymb} 
\usepackage{hyperref}
\usepackage{booktabs} 

\title{\LARGE \bf
Joint Prediction of Depths, Normals and Surface Curvature from RGB Images using CNNs 
}

\author{Thanuja Dharmasiri$^{*}$ Andrew Spek$^{*}$ Tom Drummond 
\thanks{*The authors contributed equally}
\thanks{This work was supported by the Australian Research Council Centre of Excellence for Robotic Vision (project number CE14010006). }
\thanks{The authors are with the Faculty of Electrical and Computer Systems Engineering,
        Monash University, Australia.
        {\tt\small [firstname].[lastname]@monash.edu}}%
}

\begin{document}

\maketitle

\thispagestyle{empty}
\pagestyle{empty}

\begin{abstract}

Understanding the 3D structure of a scene is of vital importance, when it comes to developing fully autonomous robots. To this end, we present a novel deep learning based framework that estimates depth, surface normals and surface curvature by only using a  single RGB image. To the best of our knowledge this is the first work to estimate surface curvature from colour using a machine learning approach. Additionally, we demonstrate that by tuning the network to infer well designed features, such as surface curvature, we can achieve improved performance at estimating depth and normals.This indicates that network guidance is still a useful aspect of designing and training a neural network. We run extensive experiments where the network is trained to infer different tasks while the model capacity is kept constant resulting in different feature maps based on the tasks at hand. We outperform the previous state-of-the-art benchmarks which  jointly estimate depths and surface normals while predicting surface curvature in parallel.


\end{abstract}


\section{INTRODUCTION}

Extracting information from raw data is a well studied problem in robotics. A visual image is one such form of raw data and has been widely used in the community to tackle a range of problems including image segmentation \cite{Long}, localization and mapping \cite{Newcombe2011}, visual servoing \cite{espiau2002new} etc. and there exist a continuous stream of research which look at maximizing the amount of information extracted. In this paper we show that we can estimate geometric quantities such as surface curvature using only RGB images as input. To our knowledge this is the first work to demonstrate such a capability. 

Surface Curvature is an important geometric surface feature, that indicates the rate at which the direction of the normals change on the surface at any particular point. It has been shown to be particularly useful for the task of segmentation on range image and 3D data \cite{Besl1988,Douros2002,Alshawabkeh2008,Lee2014}. A key challenge in accurately estimating surface curvature is its sensitivity to noise in the input data, as it is a second order surface derivative, it is affected quadratically by noise. Previous works have shown that neural networks can be used to provide accurate geometric estimates from just single RGB images \cite{Eigen,Liu,Saxena,Laina2016}, including estimating depth and normals. In this work we extend our network to estimate principal surface curvatures as well as depth and normals and demonstrate that we can accurately perform this task from a single RGB image.

\begin{figure}[t]
	\centering
	\includegraphics[width=\linewidth]{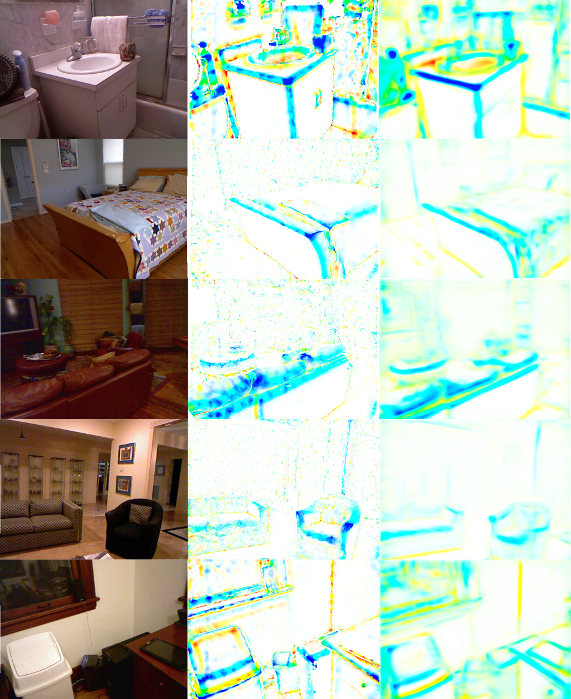} 
	\caption{\textbf{A selection of curvature predictions made by our system}. The left column shows the corresponding RGB image from the NYUv2 test dataset which was used as the only source of information to estimate curvature. The middle column shows the ground truth curvature computed using the depth data and the right image shows the prediction of our network. The Positive curvatures are shown in blue, Negatives in red, Saddles in green and Planes in white.}
	\label{fig:intro_fig}
\end{figure}

Contrary to the popular belief that hand-engineered features are inferior compared to learnt features, we argue that well designed features combined with machine learnt representations provide improved performance. It should be stressed that the features designed are not hand calculated by us, but rather predicted by the network itself as part of the inference pipeline. More concretely, we \emph{inform} the network in order to accurately estimate a single quantity such as depth, normals or curvature the network should learn an internal representation of the other two quantities. We demonstrate this by estimating surface curvature, surface normals and depth in a multi task learning framework which gives us superior results compared to training them as individual tasks. We employ a two-stage learning process where coarse level predictions of all three quantities are used as feature maps for the finer layers. Our work is similar to \cite{Eigen} in that sense, as Eigen et al. also estimated three quantities (depth, surface normals and semantic labels) using a single network. The fundamental difference between ours and their approach is that the three quantities we try to estimate are more tightly coupled at a primitive level where as the semantic labels, although clearly related, should be considered a higher order quantity compared to its counterparts depths and normals. We show both quantitatively and qualitatively that we are able to achieve better results on depth and surface normals on the NYUv2 dataset \cite{Silberman} by estimating a view point invariant quantity (surface curvature) jointly with depth and normals. We believe that robotic applications that revolve around segmentation tasks such as the Amazon Picking Challenge could benefit from our approach. Our contributions are as follows :
\begin{itemize}
	\item A novel technique to estimate surface curvature of objects using purely RGB images.(Method: Section \ref{subsec:Surface Curvature}, Results: Section \ref{subsec:Surface Curvature Results}) 
	\item A framework which predicts depth,surface normals and curvature jointly.(Method: Section \ref{sec:TASKS}, Results:  Section \ref{sec:EXPERIMENTS})
	\item Demonstrate that joint training can improve the accuracy of all three tasks while keeping the model capacity fixed (Method: Section \ref{subsec:Model Capacity}, Results: Tables \ref{tab:depth_prediction}, \ref{tab:normal_prediction}, \ref{tab:curv_prediction}.
\end{itemize}

\section{RELATED WORK}
In this section we review existing work in the literature that is related to this paper and in turn inspired the ideas presented. We take a look at traditional approaches used to compute surface curvature from raw depth data, then we summarize how deep learning has been used to predict information from images and finally, we discuss how the problem of learning multiple tasks in a single platform was performed using deep learning. 

\textbf{Surface Curvature Estimation}
Surface Curvature estimation is a well explored topic in robotics and computer vision. It has been shown to be useful for object segmentation \cite{Besl1988,Douros2002,Lee2014,Alshawabkeh2008} in depth scans and RGB-D imagery. There are several popular approaches to estimating the surface curvature. One technique is to simply twice-differentiate the surface \cite{Besl1988,Alshawabkeh2008}, but this can lead to a high sensitivity to noise in the data and generally requires removal or rejection of surface outliers. Another technique is to estimate the surface curvature from a locally connected surface mesh based on the change in adjacent facet normal angles \cite{Lee2014,Rusinkiewicza,Griffin2012}. This method is predominantly used for computer graphics and low-noise data as it operates on a small neighbourhood of facets. Yet another technique is to use locally fit surface quadrics and directly extract the principal curvatures from their parameterization \cite{Douros2002,Mitra2004}, which has been shown to be robust to noisy data and fast enough to be computed in real-time \cite{Spek2015}. In this work we use the approach in \cite{Spek2015}, to compute surface curvature and surface normals from the training data sourced from the NYUv2 dataset \cite{Silberman} as they have shown it performs well on range image data of the type present in the dataset.

\textbf{Predicting Information using Deep Neural Networks}
Convolutional Neural Networks (CNNs) have been very effectively applied to a range of robotic and vision tasks including grasp pose detection \cite{Redmon2015,Gualtieri2016}, image classification \cite{Krizhevsky,He}, semantic segmentation \cite{Long}, depth estimation \cite{Eigen,Liu,Saxena,Laina2016}, surface normal estimation \cite{wang2015designing,bansal2016marr,Eigen}. Our work is more closely related to the latter two tasks as we demonstrate surface curvature can be predicted using RGB images as the only input. We began this work by using the VGG architecture \cite{Simonyan} as a starting point to predict surface curvature in a standalone network and extended it to estimate depth and surface normals as well, in the one network. 

Prior to the resurgence of neural networks depth was either computed using a Simultaneous Localisation and Mapping (SLAM) system \cite{Klein2007,Newcombe2011} or directly obtained from a range sensor such as expensive LIDAR, stereo rigs, Time of Flight (ToF) sensors or structured light sensors\cite{newcombe2011kinectfusion} (Microsoft Kinect). In a robotic context going from data to information as efficiently as possible is vital, and predicting quantities from a single image is a step in that direction. Saxena et al. in \cite{Saxena} used a supervised learning approach that combines local and global image features by using a Markov Random Field (MRF). The idea of using both global and local features was further investigated by Eigen et al. \cite{EigenNIPS} using the AlexNet \cite{Krizhevsky} architecture in a multi-scale scheme. Liu et al. \cite{Liu} proposed to combine graphical models in the form of a Conditional Random Field (CRF) with a CNN to improve the accuracy of monocular depth estimation. More recently, Laina et al. \cite{Laina2016} proposed to use a far superior fully convolutional residual architecture and obtain state-of-the-art results in single image depth estimation. 

Data driven single image surface normal estimation was first tackled by Fouhey et al. in \cite{Fouhey}. They used a SVM based detector followed by an iterative optimization scheme to extract geometrically informative primitives. Ladicky et al. proposed to use image cues of pixel-wise and segment based methods to generate a feature representation that can estimate surface normals in a boosting framework \cite{zeisl2014discriminatively}. A ConvNet approach to estimating surface normals in global and local scales while incorporating numerous constraints such as room layout and edge labels was taken by Wang et al. \cite{wang2015designing}.  Recently, Bansal et al. \cite{bansal2016marr} showed that by combining hierarchy of features from different levels of activations in a skip-network architecture that you could generate much finer predictions for surface normals achieving state of the art results.

\textbf{Learning Multiple Tasks}
In one of the earliest works in this area Caruana et al. showed in \cite{caruana1998multitask} that by learning related tasks in parallel, the performance of all tasks could be improved, which is consistent with our findings. Multiple tasks were learned in the form of material classification and defect detection in railway fasteners in \cite{gibert2017deep} where they used Deep CNN based multi task learning for railway track inspection. They were able to show the adaptability of the multi task learning platform by using different training batch sizes (due to availability of data). In our case, all three tasks were trained with the same batch size as training data for the derived quantities (normals and curvature) were computed from depth. Multi task learning algorithms were also used to perform head pose estimation \cite{yan2016multi}, web search ranking \cite{chapelle2011boosted}, face verification \cite{wang2009boosted} etc. Li et al. in their work \emph{Learning Without Forgetting} \cite{li2016learning} demonstrated that in the presence of a model trained on one task, it can be fine-tuned to perform better on a new task while not hindering the performance of the previous task by only using training data of the new task. However, as we have access to training data for all three tasks we train the prediction stacks jointly in order to achieve superior performance compared to fine-tuning.

\section{MODEL ARCHITECTURE}

The functionalities of the model can be divided into 3 main sections. Firstly, there is a set of convolutional layers based on the VGG16\cite{Simonyan} architecture corresponding to \emph{feature extraction} which is followed by 2 fully connected layers which can see the whole image in their field views. Secondly, we have a stack of convolutional layers corresponding to \emph{coarse level predictions} of depths, normals and curvatures and finally, a set of convolutional layers which predicts the three quantities at a \emph{finer resolution}.    

\begin{figure}[h!]
	\centering
	\includegraphics[width=\linewidth, height=17cm]{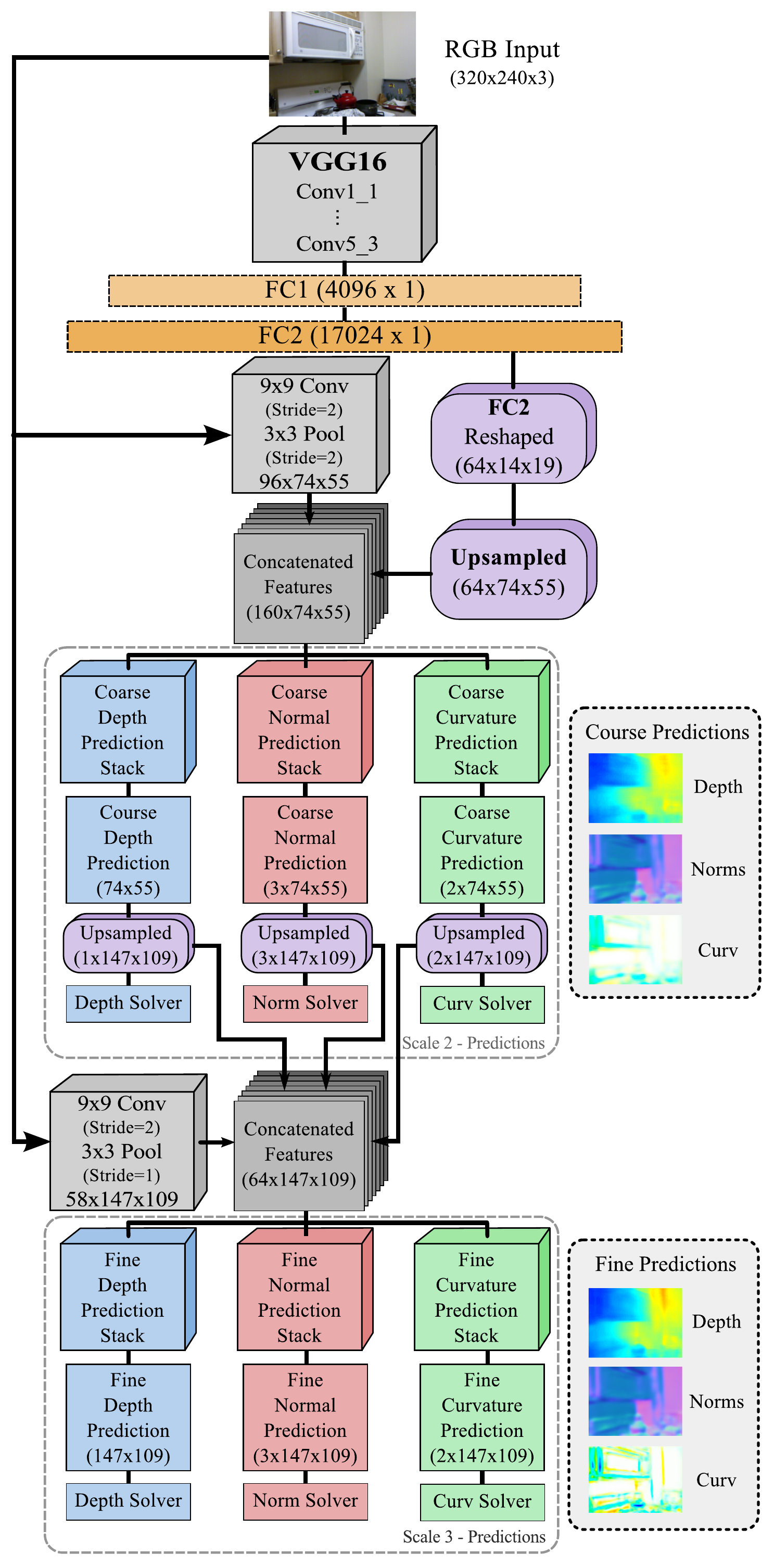}
	\captionof{figure}\textbf{{Visual Representation of Model Architecture}}
	\label{fig:Architecture}
\end{figure}

It is worth mentioning that all the convolutional layers in the coarse and fine level prediction stacks perform 5x5 convolutions with a stride of 1 and a pad of 2. Therefore, the input resolution is preserved at the output. There is an explicit up-sampling layer which up samples the coarse level prediction from 74x55 resolution to 147x109 and this is maintained throughout, by the final convolutional stack as shown in Figure \ref{fig:Architecture}. At the end of each scale, there are individual solvers for each of the training tasks which essentially compute the loss and initiate the backward propagation. 

Although the overall architecture is as explained above, in order to make sure the model capacity is kept constant and the contribution of each new task is indeed improving the previous tasks we make several changes during training which is explained in the Section \ref{subsec:Model Capacity} .

\section{TASKS}
\label{sec:TASKS}
\subsection{Depth}

We use the raw depth data distribution given by the NYUv2 dataset \cite{Silberman} for training based on the official train and test scene split (that is 249 training scenes and 219 test scenes).  Similar to previous approaches we train our network to estimate depth at several scales. The loss function used for calculating the error in the depth estimation includes an Euclidean loss term, a scale invariant term and a gradient term which compares the local rate of change of the predicted and ground truth depth values spatially. However, we do not the fix the coarse level feature stacks while training the fine level features (depth, normals and curvature) rather jointly train both stacks together as opposed to \cite{Eigen}. For the benefit of the reader we include the loss function in the following equation which is the same loss criterion employed by \cite{Eigen}.

\begin{equation}
L(D,D^*) =\sum_{i=1}^{n}d_i^2 - \frac{1}{2n^2}\left(\sum_{i=1}^{n}d_i\right)^2  + \frac{1}{n}\sum_{i=1}^{n}\left((\nabla_x d_i)^2 + (\nabla_y d_i)^2\right) 
\end{equation}
where $d_i$ is the difference in predicted log depth and ground truth log depth for the valid pixels \emph{n} (pixels that contain non-zero depth values in the raw depth data), $\nabla_x d_i$ is the horizontal image gradient of the difference and $\nabla_y d_i$ is it's vertical counterpart. 

\subsection{Surface Normals}

The ground truth normals are computed using different techniques in the  literature. We estimate the normals by fitting a quadric patch to a set of nearby points in the point cloud. This gives a more accurate representation of the surface compared to just fitting planar regions, while not adding an extra time complexity as the normals are computed as part of the curvature computation pipeline. 
We use a combination of pixel wise Euclidean loss along the three channels corresponding to the three unit vectors i, j, k and the difference in angle between the predicted normals and the ground truth as the loss criterion when the normals are trained. This is expressed as
\begin{equation}
L(N,N^*) = -N \cdot N^* + \sum_{i=1}^{3}(n_i - n^*_i)^2,  
\end{equation}
where $N$ and $N^*$ are the predicted and ground truth normal respectively, $n_i \in N$ and $n^*_i \in N^*$ are the three components ($i,j,k$) of each of the normals. Inclusion of the the Euclidean terms improves both the convergence rate and the final accuracy of the system, compared to using the dot product term alone.  

\subsection{Surface Curvature}
\label{subsec:Surface Curvature}
We use the method from \cite{Spek2015} to compute an estimate of principal surface curvatures, which is computed from a locally fit parabolic quadric. We use a sparsely sampled circular patch of radius 18 pixels, to fit a quadric at each point and extract the local principal curvature values. We limit the principal curvature $\kappa_1, \kappa_2$ to the range $ \{ -100,100 \} $ in order to avoid the estimation of implausible curvatures, effectively limiting the minimum detectable radius of curvature to be 1cm. This aligns with the precision of the system\cite{Khoshelham2012} at the distances present in the training data. This provides a dense estimate of curvature for every point (640x480), which we then bicubicly downsample to 120x160 to generate the LMDBs that can be used in the training of our network. We attempt to estimate principal curvatures directly as opposed to Gaussian or mean curvature, as we found principal curvatures to provide improved performance during training.  

We employed a Euclidean loss criterion with depth based weighting to predict surface curvature. Due to the inherent sensor noise the computed principal curvatures which are used as the ground truth tend to have a large uncertainty beyond a certain distance threshold. To prevent the network from learning these rather uncertain values we use the following loss function  
\begin{equation}
L(C,C*) = \sum_{i=1}^{n}(\frac{(\kappa_{1i} - \kappa_{1i}^*)^2 + (\kappa_{2i} - \kappa_{2i}^*)^2} {(1+D_i)^{-2}},   
\end{equation}
where $\kappa_{1i}$ and  $\kappa_{2i}$ are the predicted principal curvatures and $\kappa_{1i}^*$ and  $\kappa_{2i}^*$ are their corresponding ground truth values while D represents the depth in meters for the $i^{th}$ pixel.

\section{TRAINING}

\subsection{Data Generation}
We randomly augment the training data by performing flips, translations, rotations and variations on the color channels. The same transformation is applied to the RGB input, ground truth depth, surface curvature and surface normals in order to obtain consistent training data. Unlike some notable previous approaches \cite{Eigen}, we use the raw depth directly from the dataset provided without any post processing to fill holes or smooth surfaces. We also use the raw depths to calculate the surface normals and surface curvature, which provides a stronger link between our three ground truth sources.

As described in Section \ref{subsec:Surface Curvature} we use the method described in \cite{Spek2015} to produce training data for surface normals and surface curvature. Their approach to curvature and normal estimation is specifically targeted for noisy data such as that from a Microsoft Kinect, and they show that it produces good estimates for both surface normals and principal curvatures. We found that by scaling the ground truth curvature values by a factor of 0.1, to produce a similar range of values to the input depth, improved both qualitative and quantitative results of curvature estimation. We reverse this scaling when we provide our final prediction for both principal curvatures $\kappa_1$ and $\kappa_2$ by multiplying each value by 10.

\subsection{Hyperparameters and Weight Initialisation}
We use Nesterov’s accelerated gradient \cite{nesterov1983method} as the optimizer with a base learning rate of 0.1 and a momentum of 0.95 and train for 50 epochs using a NVIDIA GeForce GTX 1080, which took approximately 4 days. Weights of the convolutional layers corresponding to feature extraction were initialized using VGG pretrained on ILSVRC \cite{ILSVRC15} image data. We also experimented with initializing the feature extraction layers with the VGG weights of \cite{Eigen} and found that it did not give a qualitative or quantitative improvement, although it converged faster. All the convolutional layers corresponding to depth, normals and curvature estimation and the fully connected layers were randomly initialized using MSRA weight initialization scheme \cite{he2015delving} which converged much faster compared to initializing the filters from a Gaussian distribution with zero mean and 0.01 standard deviation. Everytime when the training loss plateaued (approximately every 10 epochs) we halved the learning rate and continued training. Caffe \cite{jia2014caffe} was used as the learning framework and all the experiments were carried out using a mini batch size of 16.

\begin{figure}[h!]
	\centering
	\includegraphics[width=0.7\linewidth]{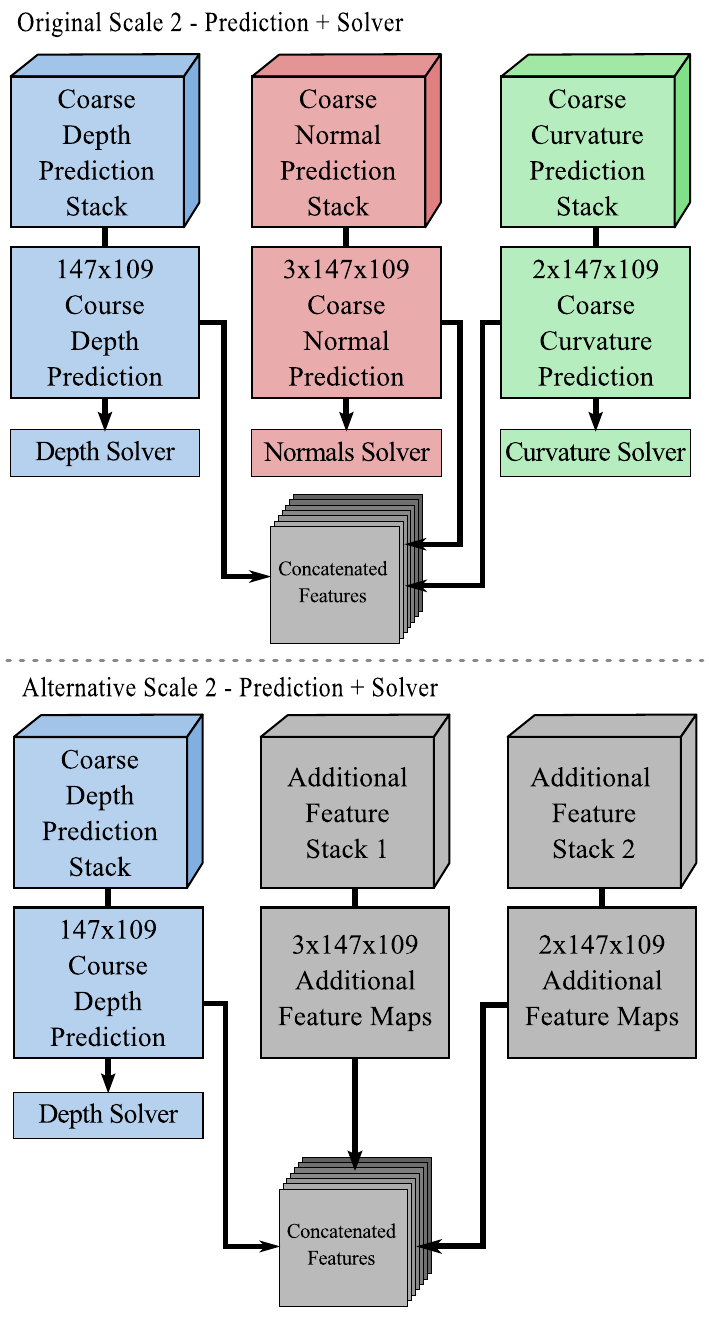}
	\captionof{figure}{\textbf{A closer look at Scale 2 of the architecture for different tasks}. \textbf{Top: } When all three tasks are trained jointly, there is a solver at the end of Scale 2 for all three tasks and the coarse feature maps are passed on to Scale 3 after being concatenated together. \textbf{Bottom: } When only a single task is trained(in this case depth) there is a single solver at the end of Scale 2 and the other two stacks now provide additional feature maps which can trained by the Scale 3 solver(not shown in the figure).}
	\label{fig:Scale2}
\end{figure}

\subsection{Training Separate Models With Equal Model Capacity}
\label{subsec:Model Capacity}
We train several models with equivalent model capacity to estimate quantities both separately and jointly. We do this to demonstrate that the improved estimates for normals and depths are not the result of increased model capacity, but more likely the result of including derived features as tasks to the network. Explicitly we train 4 models, depth only, normals only, depth+normals, depth+normals+curvature, all while maintaining a constant model capacity for each task. More concretely, when we train  a single quantity model (depths only or normals only) we leave the coarse level convolutional layers corresponding to the other tasks in place. However, there is only a single solver attached at the end of scale 2 based on the training task. We would like to point the reader to Figure \ref{fig:Scale2}, in which we are looking at the scale 2 prediction section of our model. When we are training all three tasks jointly, there is a solver attached at the end of each prediction stack. Simultaneously, we pass the coarse level predictions to scale 3 to be refined further. In a scenario where there is only one training task, the solver corresponding to the training task is kept intact while the other solvers are removed. However, the feature maps of the other stacks are still present and now act as additional weights which are trained using the scale 3 solver. To recapitulate, we preserve the model capacity by keeping the number of feature maps a constant regardless of the task/tasks that is been trained while greatly influencing what is being learnt by the feature maps through the use of additional tasks.

\section{EXPERIMENTS AND RESULTS}
\label{sec:EXPERIMENTS}
In this section we evaluate the performance of our system across the three tasks. We begin with a quantitative analysis for each of the tasks using the established benchmarks. Then we present qualitative results of our system and conclude this section with a segmentation example to showcase how this work could be applied in a real life scenario.
\begin{figure}[h!]
	\includegraphics[width=\linewidth]{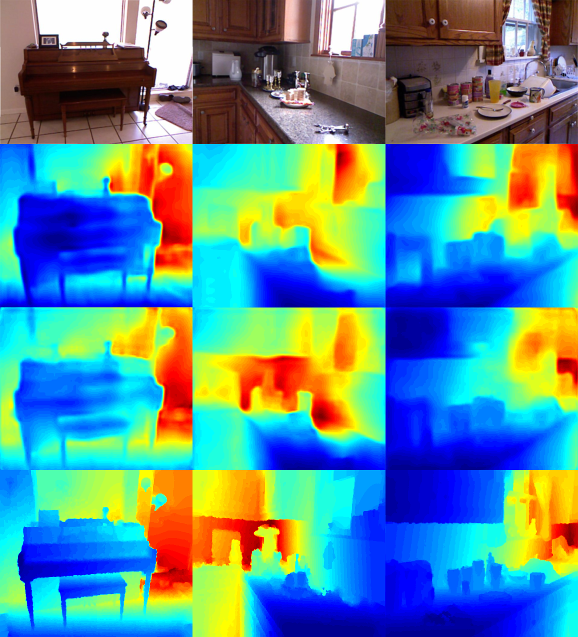}
	\caption{\textbf{Demonstrates the qualitative improvement of our approach for depth estimation}. \textbf{Top:} RGB image \textbf{1$^{\textbf{st}}$ row:} Eigen's Prediction \textbf{2$^\textbf{nd}$ row:} Our Prediction \textbf{Bottom:} Ground Truth}
	\vspace{-1em}
\end{figure}

\begin{table}[h!]
	\centering
	\resizebox{\linewidth}{!}{%
		\begin{tabular}{cccccccc}
			\multicolumn{8}{l}{\normalsize Depth Prediction} \\ \toprule
			Type &Method & $\text{Rel}_{abs}$ & $\text{RMS}_{lin}$ & $\text{RMS}_{log}$ & $\delta$ & $\delta^2$ & $\delta^3$ \\ \midrule
			\multirow{4}{*}{\rotatebox{90}{\normalsize single}}&Liu\cite{Liu} & 0.230  & 0.824 & - & 0.614 & 0.883 & 0.972 \\
			&Eigen\cite{EigenNIPS} & 0.214 & 0.877 & 0.283 & 0.614 & 0.888 & 0.972 \\
			&Ours(Depth) & 0.156  & 0.646 & 0.216 & 0.765 & 0.949 & 0.987 \\
			&Laina\cite{Laina2016} & \textbf{0.127} & \textbf{0.573} & \textbf{0.195} & \textbf{0.811} & \textbf{0.953} & \textbf{0.988} \\
			\midrule
			\multirow{4}{*}{\rotatebox{90}{\normalsize joint}}&Eigen(Alex)\cite{Eigen} & 0.198 & 0.753 & 0.255 & 0.697 & 0.912 & 0.977 \\ 
			&Ours(D+N) & \textbf{0.156}  & 0.642 & 0.215 & 0.766 & 0.949 & 0.988 \\
			&Eigen(VGG)\cite{Eigen}  & 0.158 & 0.641 & 0.214 & 0.769 & 0.950 & 0.988 \\
			&Ours(Full) & \textbf{0.156}  & \textbf{0.624} & \textbf{0.212} & \textbf{0.776} & \textbf{0.953} & \textbf{0.989} \\ 
		\end{tabular}
	}
	\caption{Depth prediction Metrics: the middle three columns indicate errors (lower better) from ground truth, the final three columns indicate the percentage of points within $\delta^n$ (higher better) of the ground truth ($\delta=1.25$). The bold values indicate the best performing method of each type (single,joint).}
	\label{tab:depth_prediction}
	\vspace{-1em}
\end{table}

\subsection{Depth}
We evaluate our depth predictions in the same manner as outlined in previous work \cite{Laina2016},\cite{Eigen} and the results are tabulated in Table \ref{tab:depth_prediction}. The predicted depth maps are upsampled by a factor of 4 to match the image resolution of 640x480 and are evaluated against the official ground truth depth maps including the filled in areas but limited to the region where there is a valid depth map projection. In terms of relative performance we improve mostly in terms of $RMS_{lin}$ and have similar performance for $Rel_{abs}$ and $RMS_{log}$, which are more related to the ratio of predicted and ground truth depths. We have included the methods that estimate depth alone as a single task for completeness, although we outperform all the methods except \cite{Laina2016} which uses a much more powerful ResNet\cite{He} architecture. Based on the results of the joint task training scheme we strongly believe that the performance of \cite{Laina2016} could still be improved had it been trained simultaneously with normals and surface curvature. 
As we keep adding more tasks that are based on related quantities we can see gains in performance. Also the contribution from curvature is much more significant (reduction of RMSE by 0.02) compared to the contribution of normals (reduction of RMSE by 0.004). As it can be seen in Table \ref{tab:depth_prediction} the contribution of semantic labels (Eigen VGG \cite{Eigen}), although very small, helps to increase the performance. But, curvature being a more tightly connected quantity to depth gives the largest improvement.   

\begin{table}[h!]
	\centering
	\resizebox{\linewidth}{!}{%
		\begin{tabular}{p{0.4cm}cccccc}
			\multicolumn{7}{l}{ Surface Normal Predictions : Compared to \cite{zeisl2014discriminatively}} \\ \toprule
			&  &\multicolumn{2}{c}{Angular Error} & \multicolumn{3}{c}{Within $t\degree$} \\
			Type  &\multicolumn{1}{c}{Method } & Mean & Median & $\leq11.25\degree$ & $\leq22.5\degree$ & $\leq30\degree$ \\ \midrule
			\multirow{4}{*}{\hspace{0.1cm} \rotatebox{90}{single}}&Ladicky \cite{zeisl2014discriminatively} & 35.3 & 31.2 & 16.4 \% &36.6\% &48.2\% \\
			&Wang \cite{wang2015designing} & 26.9 & 14.8 & 42.0\% & 61.2\% & 68.2\% \\
			&Ours (Norms) & 21.1 & 13.5 & 43.6\% & 66.6\% & 75.4\% \\
			&Bansal et al \cite{bansal2016marr} & \textbf{19.8} & \textbf{12.0} & \textbf{47.9 \%}& \textbf{70.0 \%} & \textbf{77.8 \%} \\
			\midrule
			\multirow{4}{*}{\hspace{0.1cm} \rotatebox{90}{joint}}&Eigen(Alex)\cite{Eigen} & 23.7 & 15.5 & 39.2 \% & 62.0 \% & 71.1\% \\
			&Ours (D+N) & 21.1 & 13.6 & 43.6\% &66.5\% & 75.4\% \\
			&Eigen(VGG)\cite{Eigen} & 20.9 & 13.2 & 44.4\% & 67.2\% & 75.9\% \\
			&Ours (Full) & \textbf{20.6} & \textbf{13.0} & \textbf{44.9\%} & \textbf{67.7\%} & \textbf{76.3\%} \\
			\\
			\multicolumn{6}{l}{ Surface Normal Predictions : Compared to \cite{Spek2015}} \\ \toprule
			&  &\multicolumn{2}{c}{Angular Error} & \multicolumn{3}{c}{Within $t\degree$} \\
			Type & Method & Mean & Median & $\leq11.25\degree$ & $\leq22.5\degree$ & $\leq30\degree$ \\ \midrule
			\multirow{3}{*}{\hspace{0.1cm} \rotatebox{90}{single}}&Wang \cite{wang2015designing} & 36.4 & 26.2 & 27.2\% & 45.6\% & 53.9\% \\
			&Ours (Norms) & 27.7 & 20.2 & 31.8\% & 53.7\% & 63.8\% \\
			&Bansal\cite{bansal2016marr} & \textbf{27.1} & \textbf{19.0} & \textbf{32.8}\% & \textbf{55.8}\% & \textbf{65.7}\%\\
			\midrule
			\multirow{4}{*}{\hspace{0.1cm} \rotatebox{90}{joint}}&Eigen(Alexnet)\cite{Eigen} & 29.7 & 21.8 & 30.0\% & 51.0\% & 61.0\%\\
			&Ours (D+N) & 27.7 & 20.2 & 31.7\% & 53.6\% & 63.7\% \\
			&Eigen(VGG)\cite{Eigen} & 27.3 & 19.6 & 32.3\% & {\textbf{54.7}\%} & 64.6\%\\
			&Ours (Full) & \textbf{27.2} & \textbf{19.6} & \textbf{32.9\%} & \textbf{54.7\%} & \textbf{64.7}\% \\
		\end{tabular}
	}
	\caption{The mean, median angular error and the percentage of points with an angular error less than a threshold $t\degree$ for several normal estimation approaches evaluated against two different methods \cite{zeisl2014discriminatively,Spek2015}.} 
	\label{tab:normal_prediction}
	\vspace{-1em}
\end{table}
\subsection{Surface Normals}
\begin{figure}[h!]
	\includegraphics[width=\linewidth]{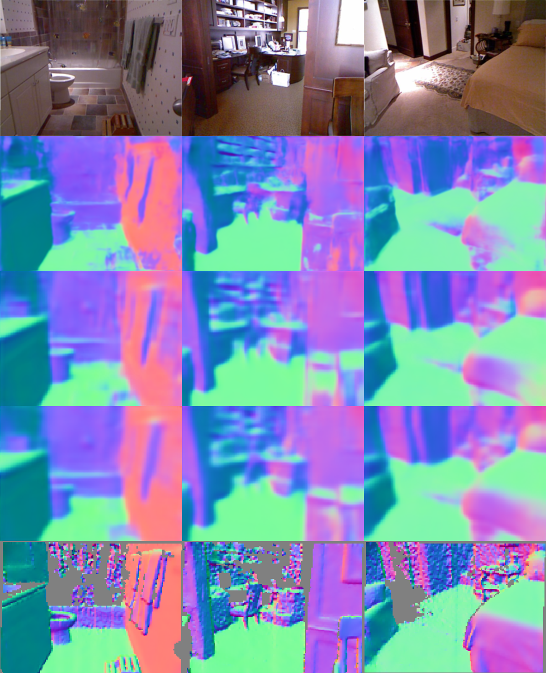}
	\caption{\textbf{Demonstrates the qualitative improvement of our approach for normal estimation}. \textbf{Top:} RGB image \textbf{1$^{\textbf{st}}$ row:} Bansal\cite{bansal2016marr},  \textbf{2$^\textbf{nd}$ row:} Eigen\cite{Eigen}, \textbf{3$^\textbf{rd}$ row:} Our Prediction \textbf{Bottom:} Ground Truth \cite{Spek2015}. The missing areas in the ground truth normals coincide with those in the raw depth images.}
	\label{fig:normal_qual}
\end{figure}

We compare our normals in a similar way to \cite{Eigen, wang2015designing, Fouhey}. As we don't have access to ground truth normal data, we compare our approach against two different methods of estimating normals from the raw depth data, including the ground truth normals as shown in \cite{zeisl2014discriminatively} and the method we use to compute our input data from \cite{Spek2015}. Qualitatively \cite{zeisl2014discriminatively} takes a more aggressive approach to noise and produces very smoothed out estimates, while the method in \cite{Spek2015} produces smoothed normals but still provides sharp edges. During evaluation the regions corresponding to the missing depth values are masked out since the ground truth normals can not be accurately computed on those areas. We summarise these results in Table \ref{tab:normal_prediction} and demonstrate improved results for each normal estimation method over previous methods. Quantitatively we approach the performance metrics of Bansal et al. \cite{bansal2016marr} who used a skip architecture with a larger model capacity compared to ours, although arguably qualitatively both \cite{Eigen} and our approach outperform their predictions as shown in Figure \ref{fig:normal_qual}.

Similar to depths, predicted normals also gained an increase in accuracy when the network was trained in a multi task platform. Although, having merely depths in parallel did not make a noticeable change extending the network to learn all three tasks resulted in a significant improvement.

\subsection{Surface Curvature}
\label{subsec:Surface Curvature Results}
In order to evaluate the accuracy of estimating surface curvature without access to true ground truth data we evaluate the performance of our approach against the method of \cite{Spek2015}. We evaluate the predictions from our network which attempts to explicitly predict curvature, to the estimated curvature values computed from the predicted depths produced by our own network and the network from \cite{Eigen}. We compare the RMS error of each of the principal curvatures ($\kappa_1, \kappa_2$) against the computed ground truth and present the median error of the \emph{mean curvature} $0.5*(\kappa_1 + \kappa_2)$ across two categories, planar and non-planar regions. We define planar surfaces to be those with a radius of curvature greater than 1 meter. As expected predicted curvatures clearly outperform the computed curvatures from depths. Furthermore, the predicted curvatures using the joint model which learned surface normals and depths in parallel provide better performance compared to the model which only learnt surface curvature. 

\begin{figure}[h!]
	\includegraphics[width=\linewidth]{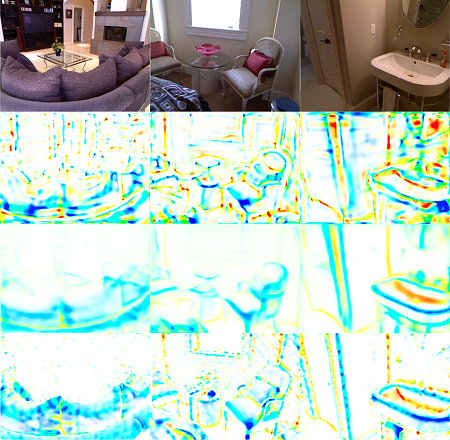} 
	\caption{\textbf{Demonstrates the qualitative improvement of our approach for surface curvature estimation}. \textbf{Top:} RGB image \textbf{1$^{\textbf{st}}$ row:} Computed surface curvature based on Eigen's\cite{Eigen} depth prediction \textbf{2$^\textbf{nd}$ row:} Prediction of our system \textbf{Bottom:} Ground Truth computed from raw depth data}
	\label{fig:curv_qual_comp}
\end{figure}

Figure \ref{fig:curv_qual_comp} is included as a reference to show how the metrics in Table \ref{tab:curv_prediction} translate into visual appearance.

\begin{table}[h]
	\centering
	\resizebox{\linewidth}{!}{ %
		\begin{tabular}{cccccccc}
			\multicolumn{8}{l}{ Principal Curvature Predictions} \\ \toprule
			&  \multicolumn{2}{c}{RMS ($m^{-1})$} &\multicolumn{2}{c}{Median ($m^{-1})$} & \multicolumn{3}{c}{Within $\sigma_t$} \\
			\multicolumn{1}{c}{Method \cite{Spek2015}} & $\kappa_1$ & $\kappa_2$ & planar & non-planar & $\sigma_1$ & $\sigma_2$ & $\sigma_3$ \\ \midrule
			Eigen(Depth) \cite{Eigen} & 5.56 & 7.50 & 3.86& 1.44& 25.7\% & 33.9\% & 43.5\% \\
			Ours (Depth) & 6.03 & 6.50 & 4.23 & 1.38& 26.9\% & 34.9\% & 44.2\% \\ 
			Ours (Curvatures) & 3.41 & 5.17&1.984 & 0.184&52.6\% & 63.2\% & 73.2\% \\
			Ours (joint) & \textbf{2.81} & \textbf{4.47}&\textbf{1.634}&\textbf{0.085} & \textbf{63.1\%} & \textbf{72.7\%} & \textbf{80.3\%} \\ 
			\\
			
		\end{tabular}
	}
	\caption{The table shows the RMS error of estimating the principal surface curvatures ($\kappa_1, \kappa_2$), the median error for planar and non-planar regions and the percentage of curvatures values that are within a threshold $\sigma_1$ = 0.25$m^{-1}$, $\sigma_2$ = 0.5$m^{-1}$, $\sigma_3$ = 1$m^{-1}$. The first two approaches do not explicitly predict curvature and are computed from the predicted depths. }
	\vspace{-1em}
	\label{tab:curv_prediction}
\end{table}

\subsection{Possible Applications For This Work}
As a purely qualitative demonstration of our approach, we show a simple example of scene segmentation that combines information from the colour, depth and curvature of selected scenes. We generate a simple segmentation by combining the gradients of colour and depth, and curvature values. This border function $b(u,v)$ can be expressed as
\begin{equation}
b(u,v) = w_I \cdot \nabla I(u,v) + w_d \cdot \nabla D(u,v) + w_c \cdot C(u,v),
\end{equation}
where $\nabla I(u,v)$ is the the magnitude of the image intensity gradient, $\nabla D(u,v)$ is the magnitude of the depth gradient and $C(u,v)$ is the curvature value at the point $u,v$. That is 
\begin{equation}
\nabla I(u,v) = \sqrt[]{\frac{\partial I(u,v)}{\partial u}^2+\frac{\partial I(u,v)}{\partial v}^2}, 
\end{equation}
and
\begin{equation}
\nabla D(u,v) = \sqrt[]{\frac{\partial D(u,v)}{\partial u}^2+\frac{\partial D(u,v)}{\partial v}^2}.
\end{equation}
The segmentation is then generated by a simple single threshold on this border function. That is a pixel is considered a border ($\mathbf{B}(u,v)$) if it satisfies the condition



\begin{equation}
\mathbf{B}(u,v) = 
\begin{cases}
1 & \text{if $b(u,v) \geq \delta_{thresh}$} \\    0 & \text{otherwise}
\end{cases}
\end{equation}

\begin{figure}[h!]
	\centering
	\includegraphics[width=\linewidth]{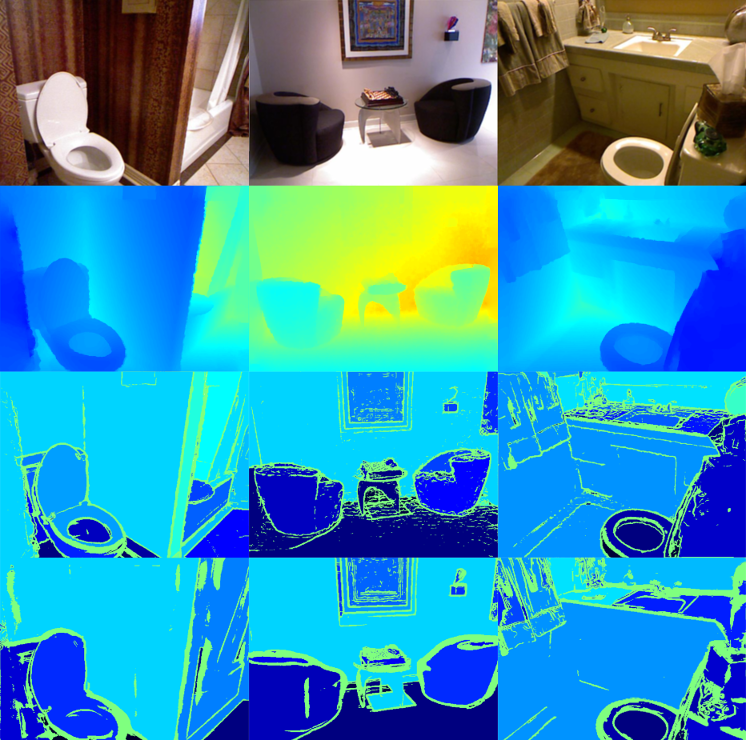}
	\caption{Demonstrates a basic segmentation algorithm, that uses colour, depth and curvature to generate a border function. The rows of the figure are, top to bottom: Input colour image, Input ground truth depth, Segmentation From GT data, Segmentation from Predicted Data. The key contribution of the depth and curvature to the segmentations, are on the depth boundaries and wall edges that are difficult to differentiate from colour alone.}
	\label{fig:segmentation_results}
\end{figure}

We compare the performance of this segmentation method using the ground truth quantities and the predictions (depths and curvature) generated by our network. We show the results of this in Figure \ref{fig:segmentation_results}. The results are not intended to be treated as state of the art segmentations, but are included to demonstrate a possible future extension of this work and also to illustrate that the information from the network can be used to perform similar tasks.

\section{Conclusions}

In this work we present a unified multi task learning platform which is capable of predicting depths, surface normals and surface curvatures using a single RGB image. We show that carefully chosen hand crafted feature representations can outperform the machine learnt features provided they are closely related to the prediction task. This shows that network guidance is a useful aspect and should not be completely ignored when training neural networks. We run extensive experiments where we keep the model capacity of the architecture fixed while gradually increasing the number of prediction tasks to verify the effectiveness of our hypothesis. We provide a potential application for our work in a robotic context in the form of a segmentation example as a qualitative demonstration.



%




\end{document}